\documentclass[letterpaper]{article} 
\usepackage{aaai2026}  
\usepackage{times}  
\usepackage{helvet}  
\usepackage{courier}  
\usepackage[hyphens]{url}  
\usepackage{graphicx} 
\usepackage{natbib}  
\usepackage{caption} 
\usepackage{algorithm}
\usepackage{algorithmic}

\usepackage{newfloat}
\usepackage{listings}
\DeclareCaptionStyle{ruled}{labelfont=normalfont,labelsep=colon,strut=off} 
\floatstyle{ruled}
\newfloat{listing}{tb}{lst}{}
\floatname{listing}{Listing}
\usepackage{amsmath}   
\usepackage{amssymb}   
\usepackage{amsfonts}  
\usepackage{physics}  
\usepackage[capitalize]{cleveref} 
\usepackage{booktabs}  

\title{Co-Layout: LLM-driven Co-optimization for Interior Layout}
\author {
    Chucheng Xiang\textsuperscript{\rm 1}, Ruchao Bao\textsuperscript{\rm 1}, Biyin Feng\textsuperscript{\rm 2},\\
	Wenzheng Wu\textsuperscript{\rm 1}, Zhongyuan Liu\textsuperscript{\rm 3}, Yirui Guan\textsuperscript{\rm 3}, Ligang Liu\textsuperscript{\rm 1}\thanks{Corresponding author.}
}
\affiliations {
    \textsuperscript{\rm 1}University of Science and Technology of China\\
	\textsuperscript{\rm 2}Tsinghua University\\
	\textsuperscript{\rm 3}Tencent\\
    \{ccxiang, iambrc, wuwzh\}@mail.ustc.edu.cn, fengby23@mails.tsinghua.edu.cn, \\\{lockliu, ezguan\}@tencent.com, lgliu@ustc.edu.cn
}

\usepackage{bibentry}

\begin{document}

\maketitle

\begin{abstract}
	We present a novel framework for automated interior design that combines large language models (LLMs) with grid-based integer programming to jointly optimize room layout and furniture placement. Given a textual prompt, the LLM-driven agent workflow extracts structured design constraints related to room configurations and furniture arrangements. These constraints are encoded into a unified grid-based representation inspired by ``Modulor". Our formulation accounts for key design requirements, including corridor connectivity, room accessibility, spatial exclusivity, and user-specified preferences. To improve computational efficiency, we adopt a coarse-to-fine optimization strategy that begins with a low-resolution grid to solve a simplified problem and guides the solution at the full resolution. Experimental results across diverse scenarios demonstrate that our joint optimization approach significantly outperforms existing two-stage design pipelines in solution quality, and achieves notable computational efficiency through the coarse-to-fine strategy.
\end{abstract}

\begin{links}
    \link{Project Page}{https://xccelephant.github.io/paper/co-layout/}
\end{links}

\section{Introduction}
Interior design significantly impacts our daily living environments. A certified interior designer must adeptly translate a client's broad requirements into detailed plans for rooms, corridors, and furniture layouts. This process requires designers' professional expertise and years of accumulated experience. Producing high-quality, detailed interior designs is both challenging and time-consuming, as it involves managing numerous design variables and balancing trade-offs between different design constraints. Developing automated design tools can significantly boost the productivity of interior designers while also stimulating their creativity.

Due to the complexity of interior design, existing approaches often separate the process into distinct phases: room layout~\cite{nauata2021house} and furniture layout~\cite{littlefair2025flairgpt}. In practice, experienced interior designers require simultaneous consideration of both elements, as furniture determines the room's functionality and the room influences the size and placement of furniture. Neglecting to co-optimize these aspects can lead to undesirable outcomes, such as insufficient space for movement.

In this work, we draw significant inspiration from the recent advancements in large language models (LLMs) \cite{wei2022emergent}, particularly their impressive performance in generative tasks. For the task of interior design, we discovered that advanced LLM agents can translate users' high-level requirements into detailed plans involving both room and furniture layouts. However, one significant limitation of LLMs is their inability to generate interior designs with precise coordinates, often resulting in conflicts within the output. To address this, we leverage LLMs to produce semantic constraints between rooms and furniture, and subsequently develop a novel optimization approach to create feasible designs that satisfy these constraints.

Finding a plausible interior layout solution that satisfies LLM-generated constraints is technically challenging. First, modeling these constraints is non-trivial, particularly regarding corridor connectivity and room accessibility. We draw on the concept of ``Modulor" from classical architecture theory \cite{corbusier2000modulor}, which discretizes space into a grid system. Each resulting cell is assigned one room label (e.g., bedroom, living room) and associated furniture label (e.g., bed, sofa) to define spatial programming. Second, the output from LLMs often includes dozens of constraints, and after discretization, the optimization involves thousands of integer variables. To accelerate this process, we propose a coarse-to-fine optimization approach. Initially, the constraints are simplified to be solved on a coarse grid with reduced design variables, and this solution then serves as a warm start for solving the problem on a finer grid.

In summary, this paper makes three contributions which have not been previously demonstrated:

\begin{itemize}
	\item We propose an automated interior design framework that simultaneously optimizes room layout and furniture placement by satisfying LLM-generated constraints.
	\item We introduce a grid-based representation to formalize LLM-derived constraints, transforming the design task into an integer programming problem.
	\item We develop a coarse-to-fine strategy to significantly accelerate the optimization process.
\end{itemize}

\begin{figure*}[t!]
    \centering
    \includegraphics[width=0.95\textwidth]{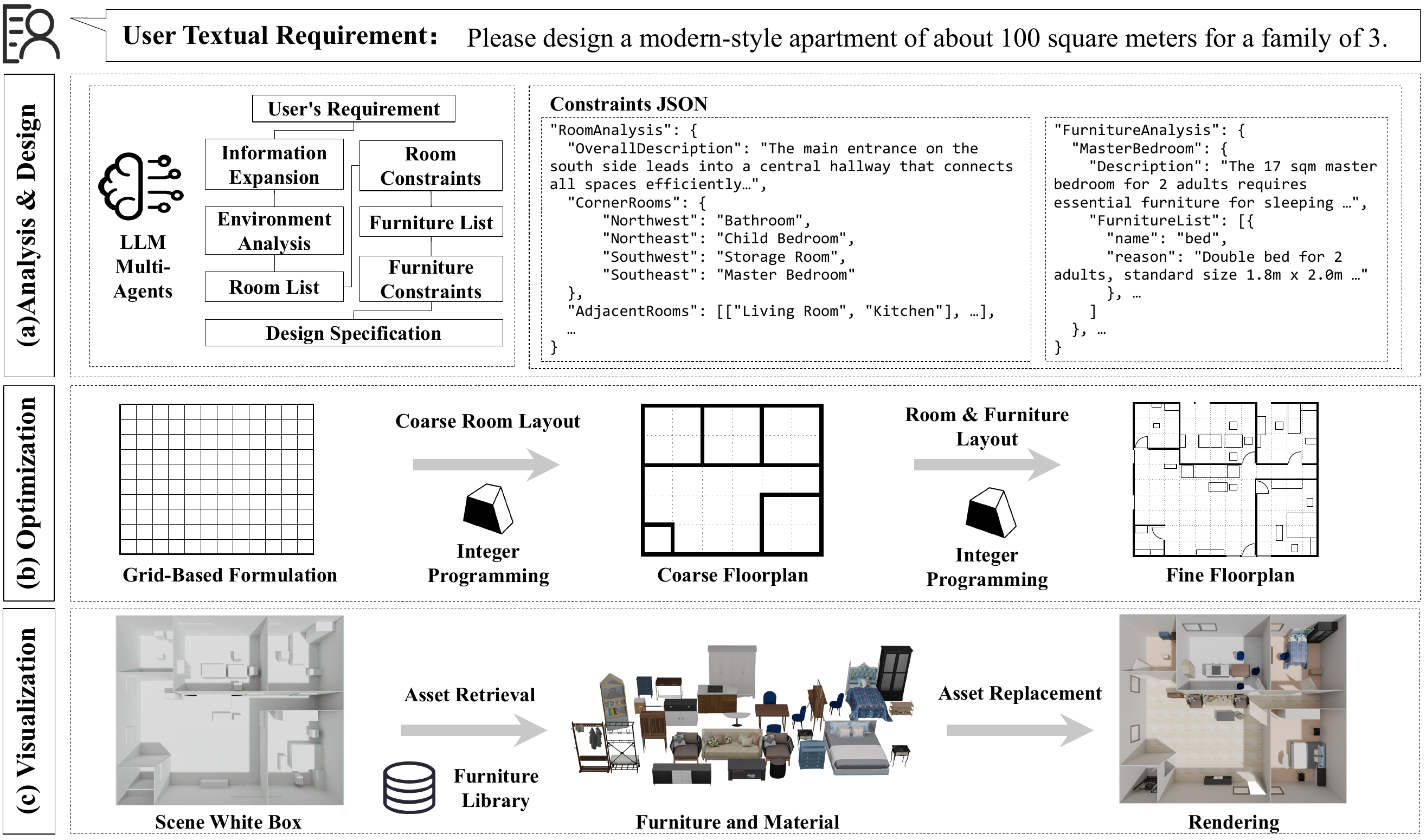}
    \caption{Overview of our automated framework. (a) First, the LLM-based workflow processes user requirements to generate spatial constraints for rooms and furniture, along with boundary conditions for the floor. (b) These constraints are then encoded into an integer programming model using our grid-based representation. The model is efficiently solved through a coarse-to-fine strategy to obtain the final layout. (c) The generated layout is converted into a scene white box in Blender~\cite{blender}. Suitable assets are then retrieved from the 3D-FUTURE~\cite{fu20213d} and Imaginarium~\cite{zhu2025imaginarium} asset libraries via semantic embedding and size matching to populate the scene.}
    \label{fig:pipeline}
\end{figure*}

\section{Related Work}
\subsubsection{Floorplan Representation.} Recent methods represent floorplans either as rasterized images or as vectorized graphs.
Rasterized approaches treat floorplans as image-like pixel matrices, and characterize rooms as segmentation masks.
Representative works include RPLAN~\cite{wu2019data}, HouseGAN~\cite{nauata2020house}, HouseGAN++~\cite{nauata2021house} and Graph2Plan~\cite{hu2020graph2plan}, etc.
These representations are limited by resolution.
On the other hand, vectorized approaches represent rooms by geometric parameters (e.g., the vertices of the room's boundary, or the position and size of the room rectangle)~\cite{wu2018miqp, para2021generative, liu2022end, shabani2023housediffusion}.
This representation better preserves geometric and topological integrity, and supports constraint modeling.
Some methods adopt pure optimization pipelines~\cite{wu2018miqp} or autoregressive generation frameworks~\cite{para2021generative} under such representations.
However, both rasterized and vectorized representations struggle to describe high-level constraints such as corridor connectivity and room accessibility in our framework.
In contrast, grid-based representation~\cite{drira2007facility, peng2014computing, hua2019integer} discretizes space into unit cells, enabling explicit modeling of such spatial constraints.
Moreover, it aligns well with the concept of ``Modulor"~\cite{corbusier2000modulor}, making it practical for real-world design scenarios.

\paragraph{Automatic Floorplan Generation.}
Floorplan generation can be categorized based on input modalities, including bubble diagrams and text descriptions.
Bubble diagrams~\cite{bubblegraph} offer geometric parameters and constraints, enabling the derivation of optimal layouts.
Textual inputs can be processed by LLMs to extract optimization settings~\cite{zong2024housellm, chen2020intelligent, qin2024chathousediffusion}.
Models can also be trained to directly map text to floorplans~\cite{leng2023tell2design}.
The generation approaches can be divided into data-driven methods~\cite{mostafavi2024floor, wu2019data, hu2020graph2plan, nauata2020house, wang2021roominoes, para2021generative, nauata2021house, sun2022wallplan, shabani2023housediffusion, hong2024cons2plan} and optimization-based techniques~\cite{wu2024free, wu2018miqp}.
However, existing methods often overlook complex constraints, such as downstream tasks like furniture placement, as well as requirements like corridor connectivity and room accessibility.
These factors are often handled manually or through post-processing~\cite{wu2018miqp}, instead of incorporating them within the generation framework.

\paragraph{Indoor Furniture Layout.}
This task typically involves selecting appropriate furniture based on input room information, and predicting their placement.
Some methods are based on optimization, such as defining constraints according to design principles and specifying optimization settings through interactions~\cite{merrell2011interactive}, using physics-based rules~\cite{weiss2018fast}, and maximizing ergonomic comfort~\cite{yu2011make}.
Some data-driven methods aim to learn spatial patterns directly from layout datasets, using models such as convolutional networks~\cite{ritchie2019fast, wang2018deep}, Transformers~\cite{paschalidou2021atiss}, or diffusion models~\cite{tang2024diffuscene, leimer2022layoutenhancer} to predict furniture arrangements.
Recent advancements utilize LLMs to interpret natural language prompts, identify relevant scene components and constraints, and guide furniture layout generation accordingly~\cite{yang2024holodeck, feng2023layoutgpt, ccelen2024design, littlefair2025flairgpt, fu2024anyhome}.
However, in interior design, furniture layout and floorplan generation are inherently interdependent. Existing approaches typically treat them as separate stages, making it difficult to achieve synergistic consideration across both levels of design.

\section{Overview}
\label{sec:overview}
Our method converts text input into a detailed interior layout design through two phases, as illustrated in \cref{fig:pipeline}.
We first introduce basic notations of a grid-based representation for floorplan modeling.
Then, an LLM-based preprocessor is proposed to transfer user input into a structural scene graph. Based on the notations and the scene graph, we formulate the joint optimization of room layout and furniture arrangement under various constraints as an integer programming problem.
Finally, we propose a coarse-to-fine strategy to solve the large-scale problem efficiently.
\begin{figure}[t]
	\centering
	\includegraphics[width=0.95\linewidth]{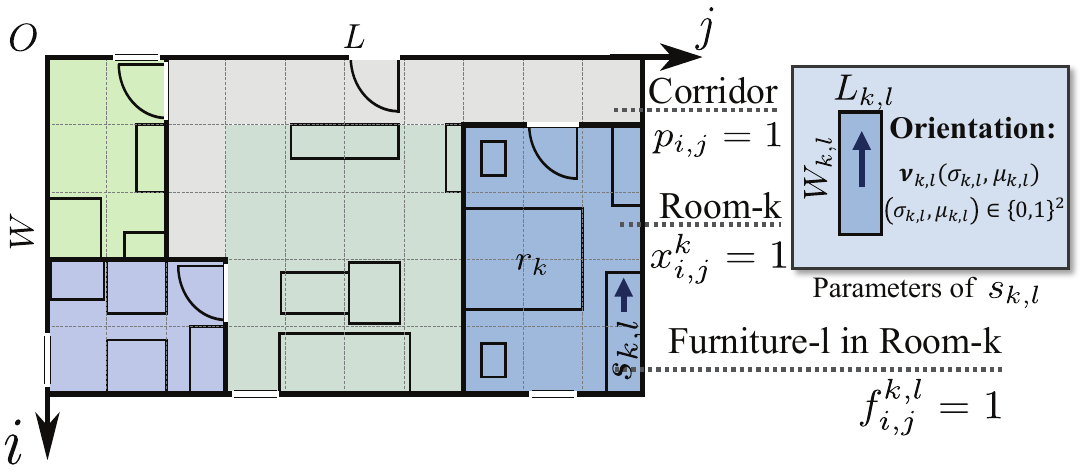}
	\caption{The floor is modeled as a 2D grid. Each cell $(i, j)$ has binary variables indicating if it belongs to a corridor ($p_{i, j}$), room $r_k$ ($x_{i, j}^k$), or is occupied by  the l-th furniture in the k-th room $s_{k, l}$ ($f_{i, j}^{k, l}$). Each furniture $s_{k, l}$ has dimensions $W_{k, l} \times L_{k, l}$. Its orientation $\boldsymbol{\nu}_{k, l}$, chosen from 4 axis-aligned directions, is set by two binary variables $\sigma_{k, l}$ and $\mu_{k, l}$.}
	\label{fig:notation}
\end{figure}
\paragraph{Notations.}
We adopt a unified grid-based representation for floorplan and furniture arrangement, see \cref{fig:notation} for an illustration.
The entire design space for a single floor is discretized into a 2D grid $\mathcal{G} = \{ (i,j) : 0 \leq i < W, 0 \leq j <L \}$, where $W$ and $L$ are the width and length of the floor rectangle (with the ``Modulor" as the basic unit).
In order to support non-rectangular boundaries, we denote the indoor space as the subset $\mathcal{G}' \subseteq \mathcal{G}$, excluding cells pre-allocated for outdoor space.
This indoor space $\mathcal{G}'$ is required to be partitioned into rooms $R = \{r_k\}_{k=1}^N$, corridors, and furniture items $S = \{s_{k, l}: 1\leq k\leq N, 1\leq l\leq N_{k}\}$ within each room.
We assign binary variables $x_{i, j}^k$, $p_{i, j}$, and $f_{i, j}^{k, l}$ to each grid cell $(i, j)$ to indicate which spatial components occupy the design space.
The orientation of the furniture $s_{k, l}$ is denoted as a vector $\boldsymbol{\nu}_{k, l}$. It only takes the four directions aligned with coordinate axes, identified as $\boldsymbol{\nu}_{k, l}(\sigma_{k, l}, \mu_{k, l}) = (\sigma_{k, l}(1 - 2\mu_{k, l}), (1 - \sigma_{k, l})(1 - 2\mu_{k, l}))$, where $(\sigma_{k, l}, \mu_{k, l})$ are two binary variables. $W_{k, l}$ and $L_{k, l}$ represent the width and length of $s_{k, l}$, which are constants determined by LLMs. For a cell $(i, j)$, its neighborhood on the grid is  $\mathcal{N}(i,j) = \{(i,j+1),(i,j-1),(i+1,j),(i-1,j)\}$.

\paragraph{LLM-based Preprocessor.}
Our optimization part takes a scene graph as input, which comprises various constraints on room layouts and furniture arrangements.
However, manually designing such a scene graph is labor-intensive.
To address this, we leverage LLMs to automate the generation of scene graphs such that users only need to provide a textual idea as input.
Specifically, we design tailored prompts for an LLM-based workflow to output structured room and furniture lists $R, S$ along with associated constraints.
In this stage, fundamental floor parameters, environmental conditions, room configurations, and furniture constraints are comprehensively considered. Further implementation details are left in the supplementary material.

\section{Formulation and Optimization}
\label{sec:formulation_optimization}
The core of our method is an IP model that co-optimizes room layout and furniture arrangement, based on interior design principles and the grid-based representation. Notably, the positions of doors and windows are not optimization variables; instead, they are determined via post-processing. And we present only the most important constraints here and refer the reader to the supplementary material for a complete formulation.

\subsection{Key Constraints}
Our optimization model organizes constraints hierarchically, from basic spatial exclusivity to complex inter-room and furniture placement rules. This structure enables the model to capture essential design interdependencies while remaining computationally efficient.

\paragraph{Non-Overlapping Constraint.}
This fundamental constraint ensures the spatial exclusivity of elements within the floor. Each cell $(i,j) \in \mathcal{G}'$ must be uniquely assigned to either a room or a corridor segment. This constraint is mathematically formulated as:
\begin{equation*}
	\label{eq:non_overlapping}
	p_{i,j} + \sum\limits_{k=1}^N x_{i,j}^k = 1, \quad \forall (i,j) \in \mathcal{G}'.
\end{equation*}
The summation guarantees mutual exclusivity and complete coverage of the assignable floor area, which underpins the optimization.

\paragraph{Corridor Connectivity Constraint.}
Open rooms are defined as rooms without walls, judged by LLMs, such as certain living rooms, which are treated as part of the corridor. To ensure that the corridor and open rooms form a single connected network, we employ a flow-based formulation. Let $\mathcal{D} = \{\text{E}, \text{W}, \text{S}, \text{N}\}$ denote the set of cardinal directions, and let $\text{adj}(i,j,d)$ return the neighboring cell of $(i,j)$ in direction $d \in \mathcal{D}$ if it exists within $\mathcal{G}'$, and $\emptyset$ otherwise. We define variables $ \text{flow}_{i,j}^d \geq 0$ representing the flow from cell $(i,j)$ in direction $d$.
Let $R_O$ denote the set of open rooms. For each cell $(i,j) \in \mathcal{G}'$, we define
\begin{equation*}
	\label{eq:q_expr}
	q_{i,j} = p_{i,j} + \sum_{k \in R_O} \left( x_{i,j}^k - \sum_{l=1}^{N_k} f_{i,j}^{k,l} \right).
\end{equation*}
The connectivity is ensured through flow capacity constraints and conservation principles. Flow can only occur between corridor or open room cells:
\begin{equation*}
	\label{eq:flow_capacity_source}
	\text{flow}_{i,j}^d \leq M \cdot q_{i,j}, \quad \forall (i,j) \in \mathcal{G}', d \in \mathcal{D},
\end{equation*}
\begin{equation*}
	\label{eq:flow_capacity_target}
	\text{flow}_{i,j}^d \leq M \cdot q_{i',j'}, \forall (i,j) \in \mathcal{G}', d \in \mathcal{D}, (i',j') = \text{adj}(i,j,d),
\end{equation*}
where $M$ is a sufficiently large constant. We define the outflow and inflow for each cell $(i,j) \in \mathcal{G}'$ as:
\begin{equation*}
	\text{outflow}_{i,j} = \sum_{d \in \mathcal{D}} \text{flow}_{i,j}^d, \quad
	\text{inflow}_{i,j} = \!\!\!\!\!\!\!\!\!\!\!\!\!\!\sum_{\substack{d \in \mathcal{D}, (i',j') \in \mathcal{G}' \\ (i',j') = \text{adj}(i,j,d^{-1})}} \!\!\!\!\!\!\!\!\!\!\!\!\!\text{flow}_{i',j'}^d
\end{equation*}
where $d^{-1}$ denotes the opposite direction of $d$ (i.e., $\text{E}^{-1}=\text{W}$, $\text{S}^{-1}=\text{N}$). The flow balance constraint below ensures that the entrance cell $(i_e, j_e)$ acts as a source supplying flow equal to the total number of corridor and open room cells, while each such cell consumes exactly one unit of flow:
\begin{equation*}
	\label{eq:flow_balance}
	\text{outflow}_{i,j} - \text{inflow}_{i,j} = \begin{cases}
		\sum\limits_{(i',j') \in \mathcal{G}'} \!\!\!\!\! q_{i',j'} & \!\!\!\! \text{if } (i,j) = (i_e, j_e),    \\
		-q_{i,j}                                                    & \!\!\!\! \text{if } (i,j) \neq (i_e, j_e).
	\end{cases}
\end{equation*}

\paragraph{Room Constraints.}
Room constraints are crucial for generating functional and geometrically sound floorplans.

\textit{Accessibility.} Each room $r_k$ must be accessible from the corridor or open rooms, ensuring that there exists a path from the entrance to each room. To guarantee unobstructed access, room cells serving as interface points must remain free of furniture placement. We introduce binary auxiliary variables $\alpha_{i,j}^k$ to indicate whether cell $(i,j)$ of room $r_k$ serves as an access point:
\begin{equation*}
	\label{eq:room_accessibility}
	\begin{aligned}
		\sum\limits_{(i,j)\in \mathcal{G}'} \alpha_{i,j}^k                                               \geq 1;
		\!\!\!\!\!\!\!\!\!\!\!\!
		\sum\limits_{(i',j') \in \mathcal{N}(i,j) \cap  \mathcal{G}'} \!\!\!\!\!\!\!\!\!\!\!\!q_{i',j'} \geq \alpha_{i,j}^k;
		\alpha_{i,j}^k                                                                                   \leq x_{i,j}^k - \sum_{l=1}^{N_k} f_{i,j}^{k,l}.
	\end{aligned}
\end{equation*}
Notably, the last constraint ensures that an access point can only be established at a room cell $(i,j)$ that belongs to room $r_k$ but is not occupied by any furniture.

\textit{Adjacency.} For specified room pairs $(r_k, r_m)$ that require spatial adjacency, we enforce the presence of at least one shared boundary. This is modeled using auxiliary binary variables $\beta_{i,j}^{k,m}$ that indicate whether cell $(i,j)$ of room $r_k$ forms an adjacency point with room $r_m$:
\begin{equation*}
	\label{eq:room_adjacency}
	\begin{aligned}
		\sum\limits_{(i,j)\in \mathcal{G}'} \beta_{i,j}^{k,m}  \geq 1;\quad \!\!\!\!\!\!\!\!\!\!\!
		\sum\limits_{(i',j') \in \mathcal{N}(i,j) \cap  \mathcal{G}'}
		\!\!\!\!\!\!\!\!\!\!\!
		x_{i',j'}^m                                            \geq \beta_{i,j}^{k,m};
		\quad
		\beta_{i,j}^{k,m}                                      \leq x_{i,j}^k.
	\end{aligned}
\end{equation*}

\textit{Bounding Box}. We employ standard big-M formulations to define the axis-aligned bounding box for each room $r_k$. Let $\gamma_{k}^{\min_i}$, $\gamma_{k}^{\min_j}$ denote the minimum coordinates (along the $i$ and $j$ axes, respectively) and $\gamma_{k}^{\text{len}_i}$, $\gamma_{k}^{\text{len}_j}$ denote the dimensions of the bounding box for room $r_k$. For all $(i,j) \in \mathcal{G}'$:
\begin{equation*}
	\label{eq:room_bbox}
	\begin{aligned}
		\gamma_{k}^{\min_i}                                 & \leq i + M \cdot(1 - x_{i,j}^k), \\
		\gamma_{k}^{\min_i} + \gamma_{k}^{\text{len}_i} - 1 & \geq i - M \cdot(1 - x_{i,j}^k), \\
		\gamma_{k}^{\min_j}                                 & \leq j + M \cdot(1 - x_{i,j}^k), \\
		\gamma_{k}^{\min_j} + \gamma_{k}^{\text{len}_j} - 1 & \geq j - M \cdot(1 - x_{i,j}^k).
	\end{aligned}
\end{equation*}
Note that these constraints do not enforce a tight fit which is encouraged later on via the room rectangularity objective.

\paragraph{Furniture Constraints.}
Furniture constraints ensure the proper arrangement of furniture within room layouts while maintaining functional rationality and spatial efficiency.

\textit{Containment}. Any grid cell $(i,j)$ occupied by the furniture $s_{k,l}$ is also part of room $r_k$:
\begin{equation*}
	\label{eq:furniture_containment_revised}
	f_{i,j}^{k,l} \leq x_{i,j}^k, \quad \forall (i,j) \in  \mathcal{G}'.
\end{equation*}

\textit{Non-Overlap}. Any grid cell $(i,j)$ can be occupied by at most one piece of furniture within the same room:
\begin{equation*}
	\label{eq:furniture_non_overlap_revised}
	\sum_{l=1}^{N_k} f_{i,j}^{k,l} \le 1, \quad \forall (i,j) \in  \mathcal{G}'.
\end{equation*}

\textit{Shape and Size}. Each furniture item $s_{k,l}$ is a rectangle with LLM-generated dimensions, which are rounded to the nearest grid cells, denoted by $\ell_{k,l}^i$ and $\ell_{k,l}^j$. We define its bounding box using minimum coordinates $\xi_{k,l}^{\min_i}, \xi_{k,l}^{\min_j}$ and dimensions $\ell_{k,l}^i, \ell_{k,l}^j$, with constraints similar to those for rooms.
In addition, the total number of cells occupied by the furniture must equal its area:
\begin{equation*}
	\sum_{(i,j) \in \mathcal{G}'} f_{i,j}^{k,l} = \ell_{k,l}^i \cdot \ell_{k,l}^j.
\end{equation*}

\begin{figure*}[t]
	\centering
	\includegraphics[width=0.95\linewidth]{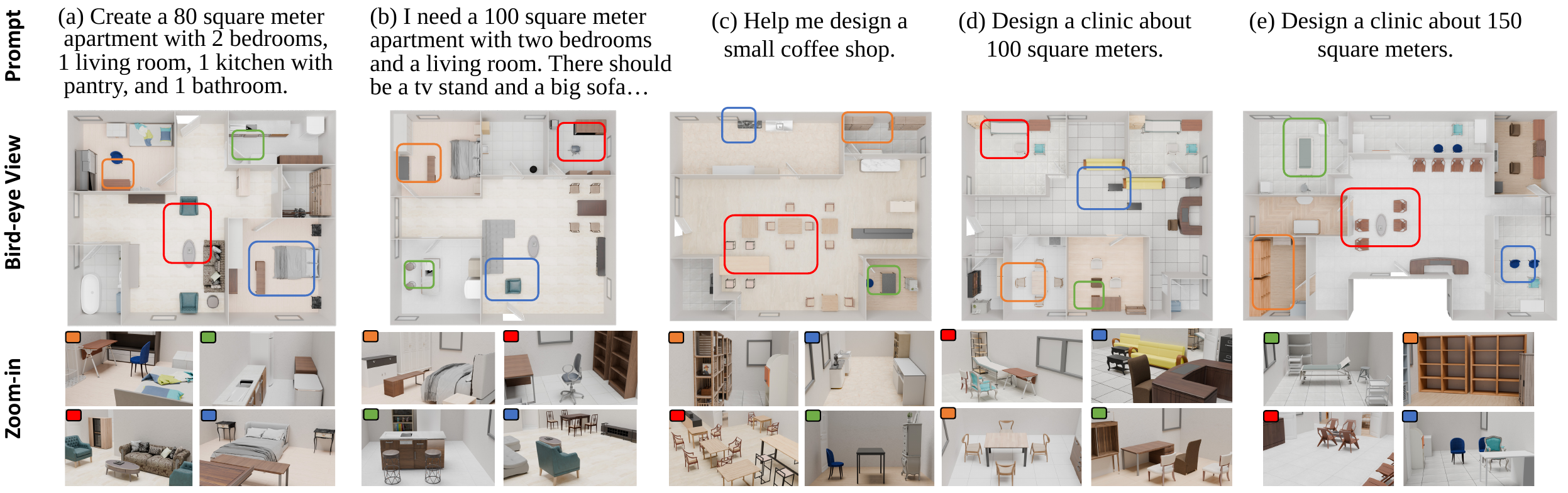}
	\caption{Various examples generated by our method, demonstrating its capability to handle diverse inputs including residential and non-residential spaces. Users can specify requirements such as building type, floor area, room functions, and required furniture. Top: Input text. Middle: Bird's-eye view of the scene. Bottom: Zoom-in views highlighting key details.}
	\label{fig:gallery}
\end{figure*}

\begin{figure*}[t]
    \centering
    \includegraphics[width=0.95\linewidth]{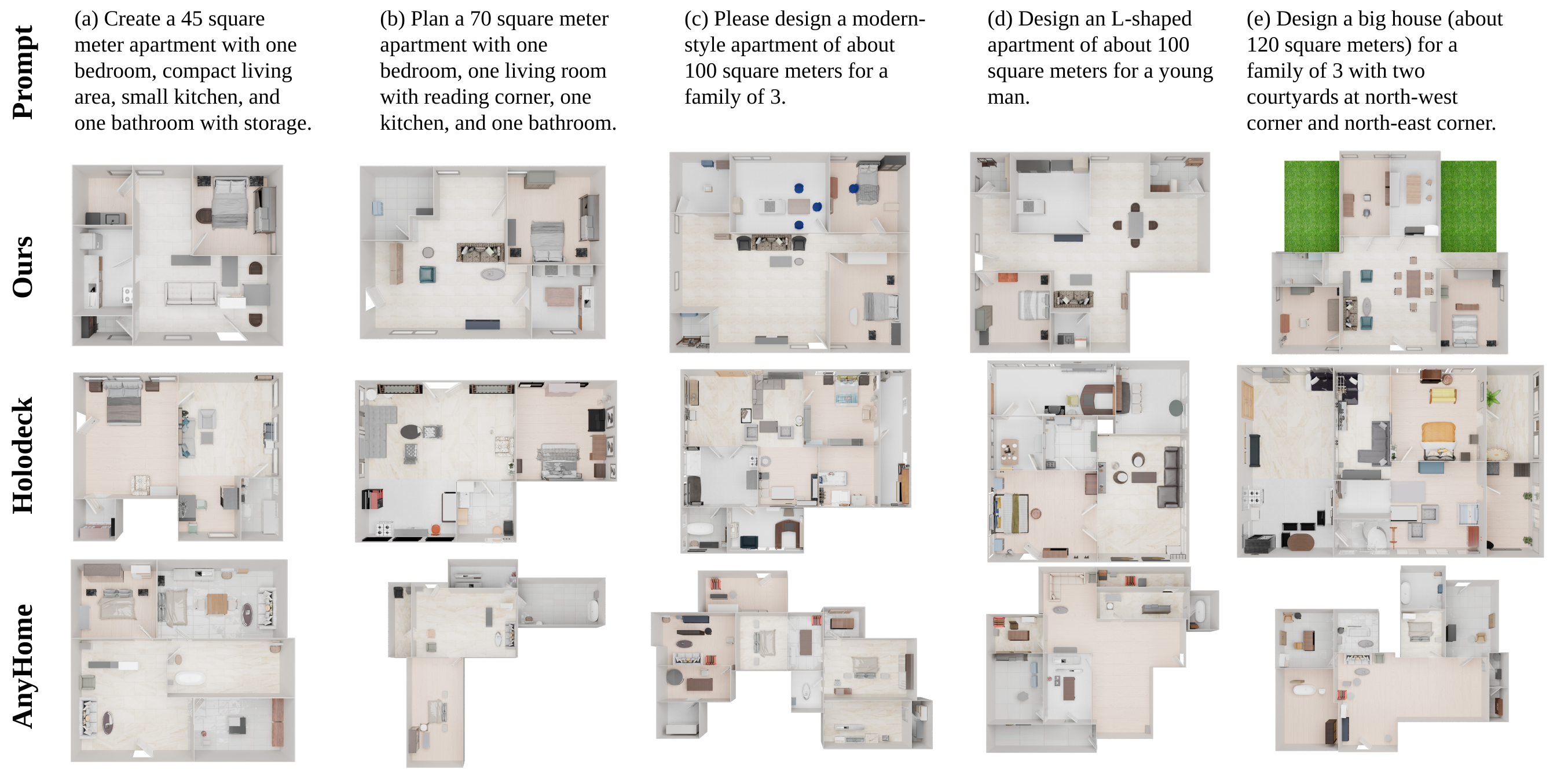}
	\caption{Comparison with baselines. Our approach consistently generates well-structured layouts with full accessibility and clear circulation. In contrast, baseline methods often produce designs with critical flaws, such as illogical circulation paths that violate privacy (e.g., Holodeck in a, d) and unreachable spaces or impractical room shapes (e.g., AnyHome in c, e).}
    \label{fig:comparison}
\end{figure*}

\subsection{Objective Functions}
\label{subsec:objective}
The overall objective function $E$ combines weighted penalty terms to guide the search toward high-quality layouts:
\begin{equation*}
    E = \sum_{s} \omega_s \cdot E_s(\mathbf{x}, \mathbf{p}, \mathbf{f}),
\end{equation*}
where $E_s$ represents individual penalty functions and $\omega_s$ denotes their respective weights. The objectives are hierarchically organized into three categories based on their impacts. We present the most important objectives here and provide a complete list in the supplementary material.

\paragraph{Geometric Quality Objectives.}
These primary objectives ensure geometrically sound and aesthetically pleasing room layouts by promoting regular and compact shapes. They also prevent rooms from being fragmented into multiple disconnected components:

\textit{Room Rectangularity}. We minimize shape irregularity by penalizing the deviation between each room's bounding box area and its actual cell count:
\begin{equation*} \label{eq:obj_rectangularity}
	E_{\text{rect}} = \sum_{k=1}^N \left(\gamma_{k}^{\text{len}_i} \cdot \gamma_{k}^{\text{len}_j} - \sum_{(i,j)\in \mathcal{G}'} x_{i,j}^k\right).
\end{equation*}

\textit{Room Perimeter}. To encourage compact room shapes, we penalize the exposed perimeter of each room:
\begin{equation*} \label{eq:obj_perimeter}
	E_{\text{perim}} = \sum_{k=1}^N \sum_{(i,j) \in \mathcal{G}'} x_{i,j}^k \cdot \abs{\mathcal{N}_{\text{out}}(i,j)},
\end{equation*}
where $\abs{\mathcal{N}_{\text{out}}(i,j)}$ denotes the number of neighboring cells not assigned to room $r_k$.

\paragraph{Room Functional Objectives.}
These objectives incorporate interior design principles and user-specified preferences to ensure functional and livable spaces.

\textit{Area Target.} Specified room area requirements are treated as soft constraints, so we penalize deviations from target areas $A_k^{\text{target}}$:
\begin{equation*} \label{eq:obj_area}
	E_{\text{area}} = \sum_{k=1}^N \abs{A_k^{\text{target}} - \sum_{(i,j) \in \mathcal{G}'} x_{i,j}^k}.
\end{equation*}

\textit{Aspect Ratio Control}. Extreme room proportions are discouraged through minimizing the difference between room length and width:
\begin{equation*} \label{eq:obj_aspect}
	E_{\text{aspect}} = \sum_{k=1}^N \abs{\gamma_{k}^{\text{len}_i} - \gamma_{k}^{\text{len}_j}}.
\end{equation*}

\paragraph{Furniture Placement Objectives.}
These objectives ensure functional and rational furniture arrangements by governing spatial relationships and distribution balance.

\textit{Spatial Relationships}. For furniture pairs with relative positioning requirements, we minimize deviation from target arrangements:
\begin{equation*}\label{eq:obj_furniture_rel}
	E_{\text{rel}} = \sum_{k=1}^N \sum_{l_1,l_2 \in \mathcal{A}^k_{\text{rel}}} \left\| \mathbf{c}_{k,l_1} - \mathbf{c}_{k,l_2} - \boldsymbol{\delta}_{l_1,l_2} \right\|_1,
\end{equation*}
where $\mathcal{A}^k_{\text{rel}}$ denotes the set of furniture pairs with specified relative positions in room $r_k$, $\mathbf{c}_{k,l}$ denotes the centroid of furniture $s_{k,l}$ and $\boldsymbol{\delta}_{l_1,l_2}$ is the target offset vector between the furniture centroids.

\textit{Distribution Balance}. To promote balanced furniture layouts, we minimize the distance between the area-weighted furniture centroid and room geometric centroid:
\begin{equation*}\label{eq:obj_furniture_balance}
	E_{\text{bal}} = \sum_{k=1}^N \left\| \hat{\mathbf{C}}_k - \mathbf{C}_k \right\|_1,
\end{equation*}
where $\hat{\mathbf{C}}_k$ is the area-weighted centroid of all furniture in room $r_k$ and $\mathbf{C}_k$ is the room's geometric centroid.

\subsection{Coarse-to-Fine Optimization Strategy}
\label{subsec:coarse_to_fine}

The joint room-furniture optimization exhibits exponential computational complexity with respect to grid resolution, making large scale instances computationally intractable. To address this challenge, we utilize a coarse-to-fine strategy.

\paragraph{Coarse Phase.}
The coarse-level formulation leverages spatial down-sampling via grid aggregation. For example, when down-sampling from a 12$\times$10 grid to a 6$\times$5 grid, each coarse cell represents a 2$\times$2 block of fine cells, reducing the problem scale by 75\%. At this level, we focus solely on room layout, deliberately deferring furniture placement to facilitate efficient exploration of high-level spatial organization.

\paragraph{Coarse to Fine Mapping.}
We need to transfer the coarse solution back to the original fine-grained grid. For each fine-level grid cell $(i, j)$, we determine its corresponding coarse-level grid cell $(i', j')$. Assuming a down-sampling factor of $s_i$ along the $i$-axis and $s_j$ along the $j$-axis, the mapping can be defined as:
\begin{equation*}
	(i', j') = \left( \left\lfloor \frac{i-1}{s_i} \right\rfloor + 1, \left\lfloor \frac{j-1}{s_j} \right\rfloor + 1 \right), \quad \forall (i, j) \in \mathcal{G}_{\text{fine}},
\end{equation*}
where $\mathcal{G}_{\text{fine}}$ is the set of fine grid cells. If coarse cell $(i', j')$ is assigned to room $r_k$ (i.e., $x_{i',j'}^k = 1$), the corresponding fine cell $(i, j)$ is provisionally assigned to the same room.

\paragraph{Fine Phase.}
This stage performs the full optimization on the original grid, incorporating both room planning and furniture placement variables.
Firstly, the mapped coarse solution provides a \textit{warm-start initialization} for the room assignment variables $x_{i,j}^k$ on the fine grid. However, note that this warm-start is not always feasible; in some cases, it may lead to infeasible solutions. Therefore, we introduce an additional mechanism.
Secondly, we add penalty terms to the fine-level objective function to encourage consistency with the coarse-level room assignments. Specifically, for each fine grid cell $(i,j)$, we penalize if it is not assigned to the same room $k=\pi(i,j)$ (the coarse assignment) by:
\begin{equation*}
	E_{\text{ref}} = \sum_{(i,j) \in \mathcal{G}_{\text{fine}}} (1 - x_{i,j}^{\pi(i,j)}) \cdot z_{i,j},
\end{equation*}
where binary variable $z_{i,j}$ indicates whether the corresponding coarse cell was assigned to any room or not.

\section{Experiments}
\subsection{Experiment Settings}
\paragraph{Environment.} We implement our algorithm in Python, using the integer programming solver GUROBI~\cite{gurobi}, version 12.
Experiments were mainly conducted on a laptop equipped with an 8-core Ryzen processor and 16 GB of RAM.
\paragraph{Metrics.} We assess the generated floorplans in three key dimensions: physical plausibility, image quality/aesthetics, and text-image alignment.
\begin{itemize}
	\item Physical Plausibility: Measured by the object overlap rate (OOR) and out of boundary rate (OOB). OOR quantifies the percentage of overlapping area between objects, while OOB calculates the proportion of objects extending beyond room boundaries.
	\item Image Quality \& Aesthetics: Evaluated using the Q-Align model~\cite{wu2023q} for both image quality assessment (IQA) and image aesthetic assessment (IAA).
	\item Text-Image Alignment: Assessed by the CLIP~\cite{radford2021learning} similarity score between the input text and the rendered floorplan image.
\end{itemize}

\paragraph{Baselines.} We compare our co-optimization approach with state-of-the-art two-stage baselines—Holodeck~\cite{yang2024holodeck} and AnyHome~\cite{fu2024anyhome}—which sequentially generate room layouts before populating furniture. For fair comparison, we use their methods solely to generate layout information, convert the results into a unified format, and render them using identical visualization scripts.

\subsection{Quantitative Evaluations}
The quantitative comparison based on our three evaluation metrics is presented in \cref{tab:quantitative_evaluations}. The evaluation is conducted on the five examples shown in \cref{fig:comparison}. Our method achieves superior performance across all metrics, particularly eliminating object overlaps and boundary violations.


\begin{table}[htbp]
	\centering
	\begin{tabular}{lccccc}
		\toprule
		Method   & OOR$\downarrow$ & OOB$\downarrow$ & IQA$\uparrow$ & IAA$\uparrow$ & CLIP$\uparrow$ \\
		\midrule
		Holodeck & 0.82            & 2.33            & 4.03          & 3.32          & 25.15          \\
		AnyHome  & 0.00            & 0.04            & 4.10          & 3.32          & 25.75          \\
		Ours     & \textbf{0.00}   & \textbf{0.00}   & \textbf{4.17} & \textbf{3.35} & \textbf{26.50} \\
		\bottomrule
	\end{tabular}
	\caption{Quantitative evaluation results. Our method achieves the best performance in terms of physical plausibility, image quality/aesthetics, and text-image alignment.}
	\label{tab:quantitative_evaluations}
\end{table}

\subsection{Qualitative Evaluations}
\paragraph{Method Demonstration.}
We present representative results of our method in \cref{fig:gallery}, demonstrating the open-vocabulary generation capability.
The gallery begins with two apartments (a) and (b), showcasing our method's capability in generating residential layouts with detailed requirements. Then, a small coffee shop (c) showcases adaptability to commercial scenarios. A 100$\mathrm{m}^2$ clinic (d) demonstrates the method's capability in professional spaces. Finally, a 150$\mathrm{m}^2$ clinic (e) further validates the generation ability for larger professional spaces. These diverse examples highlight the flexibility of our approach in handling various inputs.

\paragraph{Comparative Analysis.}
We compare our method with Holodeck and AnyHome in \cref{fig:comparison}. While baseline methods can generate a rich variety of rooms and furniture, their layouts often exhibit randomness, indicating a lack of robust understanding of room function, privacy, and connectivity. In contrast, our co-optimization approach consistently produces well-structured designs with logical spatial hierarchy and clear circulation paths, where every room is connected to a common corridor. The baselines frequently violate this principle; for instance, Holodeck's layouts may require passing through a bedroom to reach the main living area (a, d), while AnyHome can generate entirely unreachable rooms (c, e). Furthermore, our method ensures practical room shapes, whereas AnyHome often generates irregular or narrow rooms, and Holodeck's results lack coherent circulation. In summary, our joint optimization of rooms and furniture ensures the resulting layouts are not only geometrically sound and fully accessible but also functionally coherent, demonstrating a deeper understanding of architectural logic.

\subsection{User Study}
To evaluate the quality of the generated floorplans, we conducted a user study with 71 participants. After filtering invalid submissions (e.g., completion times under two minutes or uniform high/low scores), we retained 64 valid responses for analysis.
The study involved two tasks.
First, participants rated the floorplans shown in \cref{fig:comparison} on a 5-point scale based on three criteria: semantic alignment, layout rationality, and path clearance.
Second, for each prompt, participants ranked the outputs from our method and the baselines. We calculated the average scores for the rating criteria and the Mean Reciprocal Rank (MRR) \cite{voorhees1999trec} for the ranking task.
The aggregated results are presented in \cref{tab:user_study}, which shows that our method achieved the highest scores across all metrics.
\begin{table}[t]
    \centering
    \begin{tabular}{lcccc}
        \toprule
		Method & Semantic.$\uparrow$ & Layout.$\uparrow$ & Path.$\uparrow$ & MRR.$\uparrow$ \\
	  \midrule
        Holodeck & 3.43 & 3.12 & 3.06 & 0.59 \\
        AnyHome & 3.07 & 2.59 & 2.80 & 0.45 \\
        Ours & \textbf{3.77} & \textbf{3.23} & \textbf{3.41} & \textbf{0.80}\\
	  \bottomrule
    \end{tabular}
    \caption{User study results. Our method achieved the highest scores across all metrics.}
    \label{tab:user_study}
\end{table}

\begin{figure}[htbp]
    \centering
    \includegraphics[width=0.95\linewidth]{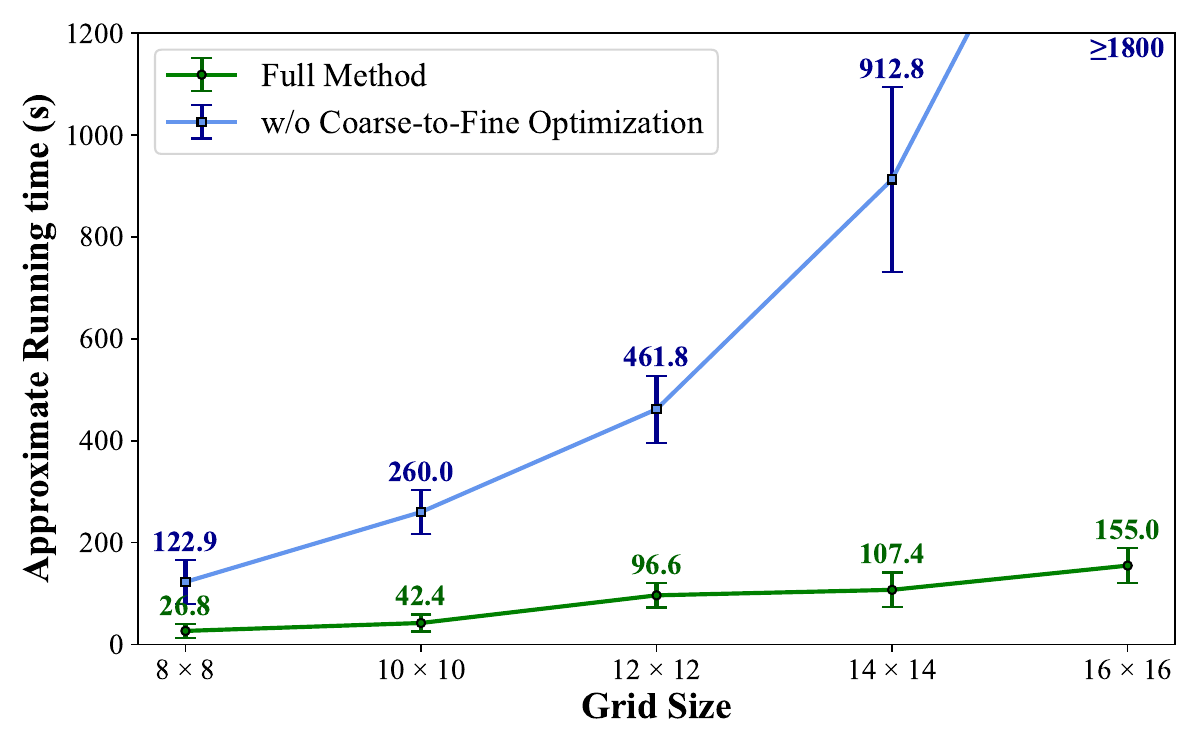}
    \caption{Ablation result of the coarse-to-fine strategy.}
    \label{fig:ablation_coarse_to_fine}
\end{figure}

\subsection{Ablation Study}
\paragraph{Ablation of Coarse-to-fine Strategy.}
To evaluate the effectiveness of our coarse-to-fine strategy, we conducted an ablation study on its impact on computational efficiency. We tested five times with a fixed prompt, varied the grid resolution, and calculated the mean and standard deviation of the running time required to obtain a feasible solution with a similar optimization gap. As illustrated in \cref{fig:ablation_coarse_to_fine}, the results show that our strategy significantly reduces computation time as the grid resolution increases, while maintaining comparable solution quality. Additional ablation studies on key constraints and the co-optimization approach are detailed in the supplementary material.

\section{Conclusion}
We present a novel framework that integrates LLMs with integer programming to automate interior layout design, jointly optimizing room layout and furniture placement. By leveraging LLMs to extract structured scene graphs from high-level user input and formulating the design problem on a grid-based representation, our method ensures corridor connectivity, room accessibility, and spatial coherence, outperforming two-stage pipelines in various metrics. A coarse-to-fine strategy further enhances computational efficiency.

Our work still holds limitations that open avenues for future research.
First, the scope of furniture is currently limited to floor-based objects. Our framework could be extended to support wall-mounted items and tabletop objects.
Second, LLMs may occasionally produce conflicting constraints, degrading solution quality. Iterative refinement or human-in-the-loop feedback could mitigate such situations.
Finally, our framework could be extended to design multi-story buildings, by incorporating inter-floor constraints.

\section*{Acknowledgments}
We thank Dr. Ziqi Wang from The Hong Kong University of Science and Technology for his valuable guidance and advice on this work.
We also thank the anonymous reviewers for their appreciation of our work. This work is supported by the National Natural Science Foundation of China (No. 62025207).

\bibliography{refer}

\clearpage
\appendix

\twocolumn[
  \begin{center}
    \LARGE \bf Supplementary Material
  \end{center}
  \vspace*{1em}
]
This supplementary material provides additional details to complement our main paper. We first elaborate on our multi-agent LLM workflow, which translates high-level user requirements into structured design specifications. We then present the detailed formulation and optimization, including additional constraints for room and furniture, and additional objectives for layout quality. We also detail the heuristic-based post-processing steps for generating architectural elements like walls, doors, and windows. Furthermore, we provide an illustrative diagram of our coarse-to-fine optimization strategy, detailed per-case results from our user study, and additional ablation studies that validates the effectiveness of key constraints and the co-optimization approach.

\section{LLM Agents for Interior Design}
\label{sec:llm_analysis}
Due to the world knowledge and reasoning capabilities~\cite{chang2024survey}, LLMs have been adopted for design tasks in various domains.
To enhance LLMs’ design capabilities, researchers have explored numerous methods, such as prompt engineering~\cite{zamfirescu2023johnny, wei2022chain} and multi-agent systems~\cite{guo2024large, hong2023metagpt}. To transform user needs into a structured specification suitable for optimization, we design a multi-agent LLM system, simulating an architect's design workflow, bridging the gap between textual descriptions and interior design constraints. This workflow involves specialized agents for:

\paragraph{Basic Information Analysis.} This agent processes the initial user requirements to extract fundamental floor parameters, including floor envelope (length, width), and floor height. It also analyzes user lifestyle preferences and analyzes family structure as critical inputs for subsequent design stages.

\paragraph{Environment Analysis.} This agent evaluates the environmental factors, including climate conditions (temperature, wind direction, sunlight, precipitation), and orientation advantages for the four cardinal directions (north, south, east, west). The analysis results in specific recommendations for each room type, identifying suitable orientations based on their functional requirements, enhancing the building's performance and occupant comfort.

\paragraph{Specific Space Analysis.}
A collection of specialized agents handle positions for critical architectural elements: \textit{Outdoor Space Analysis Agent} defines regions within the floor envelope that do not belong to indoor space. These include user-requested features such as courtyards or patios, as well as setbacks or voids necessary to achieve specific architectural forms (e.g., L-shaped buildings). \textit{Entrance Analysis Agent} determines the optimal location of the main entrance by analyzing indoor-outdoor connections and environmental conditions. The agent selects a specific point along the floor boundary that best balances these considerations.

\paragraph{Room Analysis.} This agent defines the list of required rooms. For each room, it specifies not only the desired area based on the furniture demanded to be placed inside but also detailed attributes such as key activities, performance needs (quietness, natural light), and suggested orientation. Furthermore, the agent establishes the spatial relationships of rooms. These include desired adjacencies (e.g., kitchen near dining room), and specific placement guidance like corner allocations.

\paragraph{Furniture Analysis.} This agent is responsible for populating rooms with appropriate furniture. For each room, the agent first lists essential furniture based on the functionality and then determines suitable dimensions for each piece, considering the room's area. While this agent does not perform the final detailed placement, it also identifies explicit user preferences and common-sense constraints for furniture arrangement. For instance, it might specify that a chair should face to a desk. These identified furniture types, dimensions, and constraints contribute to the structured design specification that inputs into the optimization model.

These LLM agents are guided by tailored prompts that incorporate design principles, enabling them to interpret user requirements effectively. For conciseness, the prompt for each agent is defined in a separate file. The final output of this stage is a structured design specification:
\begin{itemize}
	\item overall floor information and spatial analysis;
	\item a list of rooms with attributes;
	\item a list of furniture with dimensions for each room;
	\item spatial constraints of rooms and furniture.
\end{itemize}

\section{Details for Formulation and Optimization}
In this section, we provide some omitted details of our optimization model.

\subsection{Additional Room Constraints}
\textit{Corner Positioning}. For rooms requiring specific corner placement (e.g., a master bedroom in the northeast corner), we constrain the corresponding corner grid cell to be occupied by that room. Let $\mathcal{C} = \{\text{NW}, \text{NE}, \text{SW}, \text{SE}\}$ be the set of corner types. Each corner $c \in \mathcal{C}$ corresponds to a specific grid cell coordinate $(i_c, j_c)$. For a room $r_k$ with a specified corner preference $c_k \in \mathcal{C}$, we enforce that the cell $(i_{c_k}, j_{c_k})$ must be part of room $r_k$:
\begin{equation*}
    \label{eq:constraint_corner}
	x_{i_{c_k}, j_{c_k}}^k = 1.
\end{equation*}

\subsection{Additional Furniture Constraints}

\textit{Against Wall}. User preferences or common design principles, often derived from the LLM preprocessor, can require placing a specific furniture item $s_{k,l}$ against a wall. We model this by ensuring that at least one grid cell occupied by the furniture is adjacent to a cell outside of its assigned room $r_k$. This constraint is similar to the \textit{room adjacency} constraint. Here, we introduce auxiliary binary variables $\phi_{i,j}^{k,l}$ to indicate whether a cell $(i,j)$ occupied by furniture $s_{k,l}$ is adjacent to a cell not belonging to room $r_k$:
\begin{equation*}
    \label{eq:furniture_against_wall}
    \begin{aligned}
        \sum\limits_{(i,j)\in \mathcal{G}'} \phi_{i,j}^{k,l} \geq 1; \quad \!\!\!\!\!\!\!\!\!\!\!
        \sum\limits_{(i',j') \in \mathcal{N}(i,j) \cap  \mathcal{G}'} \!\!\!\!\!\!\!\!\!\! (1-x_{i',j'}^k) \geq \phi_{i,j}^{k,l}; \ \ 
        \phi_{i,j}^{k,l} \leq f_{i,j}^{k,l}.
    \end{aligned}
\end{equation*}
Moreover, we need an orientation constraint to ensure that objects do not face the wall. For all $(i,j) \in \mathcal{G}'$:
\begin{equation*}
	f_{i, j}^{k, l} \leq x_{i',j'}^k,
\end{equation*}
where $(i',j') = (i,j)+\boldsymbol{\nu}_{k, l}$ represents the grid cell in front of $(i, j)$.

\textit{Alignment}. For furniture pairs $(s_{k,l_1}, s_{k,l_2})$ requiring consistent orientation alignment, they should be oriented parallel to each other:
\begin{equation*}
	\sigma_{k, l_1} = \sigma_{k, l_2}.
\end{equation*}

\textit{Facing}. In addition, certain pieces of furniture may need to be positioned in a specific orientation, such as a chair $s_{k, l_1}$ facing a table $s_{k, l_2}$. This constraint can be defined as:
\begin{equation*}
	\boldsymbol{\nu}_{k, l_1} \cdot \Delta\boldsymbol{\xi} \geq 0,
\end{equation*}
where $\Delta\boldsymbol{\xi} = (\xi_{k,l_2}^{\min_i} - \xi_{k,l_1}^{\min_i}, \xi_{k,l_2}^{\min_j} - \xi_{k,l_1}^{\min_j})$ represents the direction vector pointing from $s_{k, l_1}$ to $s_{k, l_2}$ and `$\cdot$' represents standard inner product.

\subsection{Additional Objectives and Weights Settings}

\textit{Privacy Sorting}. For rooms requiring privacy ordering (e.g., bedrooms more private than living areas), we enforce a distance-based hierarchy from the entrance using soft constraints. Let $D_k$ represent the average Manhattan distance from room $r_k$ to the entrance:
\begin{equation*}
	D_k = \frac{1}{A_k^{\text{target}}} \sum_{(i,j) \in \mathcal{G}'} x_{i,j}^k \cdot d_{i,j},
\end{equation*}
where $d_{i,j} = |i - i_e| + |j - j_e|$ denotes the Manhattan distance from cell $(i,j)$ to the entrance $(i_e, j_e)$, and $A_k^{\text{target}}$ is the target area of room $r_k$. For a privacy ordering sequence $\mathcal{O} = \{k_1, k_2, \ldots, k_m\}$ where rooms are arranged in decreasing order of privacy requirements, we introduce slack variables $\theta_i \geq 0$ and enforce:
\begin{equation*}
	D_{k_i} + \theta_i \geq D_{k_{i+1}}, \quad \forall i \in \{1, 2, \ldots, m-1\}.
\end{equation*}
The privacy sorting objective penalizes violations of the expected privacy hierarchy:
\begin{equation*} \label{eq:obj_privacy}
	E_{\text{priv}} = \sum_{i=1}^{m-1} \theta_i.
\end{equation*}

\paragraph{Weights Settings.}
The relative importance of each objective is controlled by weight parameters $\omega_s$. These weights are configured based on design priorities and refined through empirical trials to ensure a suitable balance among all objectives. The specific values used in our implementation are detailed in a separate JSON file, `config.json'.

\begin{figure}[t]
	\centering
	\includegraphics[width=0.95\linewidth]{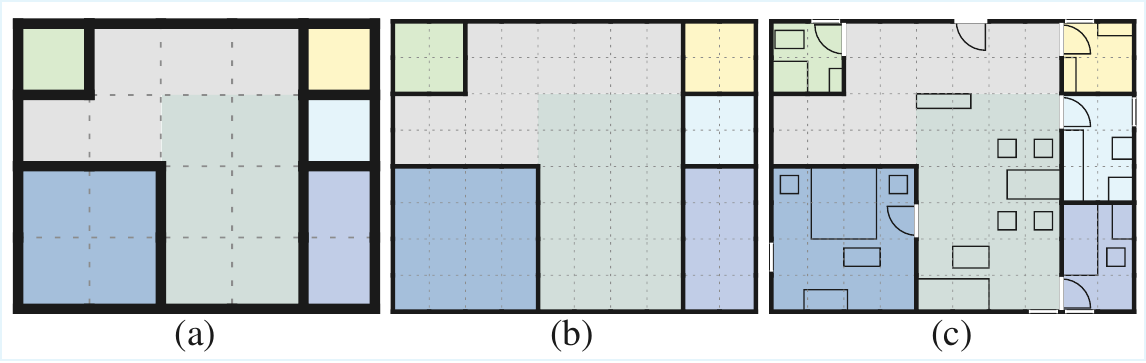}
	\caption{Illustration of the coarse-to-fine strategy for floorplan generation. (a) Resulting rooms after optimization on a coarse grid. (b) Spatial interpolation process mapping the coarse grid solution to the original fine grid. (c) Final integrated layout after joint optimization for rooms and furniture on the fine grid, guided by the interpolated solution.}
	\label{fig:coarse_fine_example}
\end{figure}

\subsection{Illustration of Coarse-to-Fine Strategy}
To visually illustrate our proposed coarse-to-fine optimization strategy, we provide a schematic diagram in \cref{fig:coarse_fine_example}.

\subsection{Post-processing Steps}
After the optimization phase, we perform several post-processing steps to generate the final architectural elements, including walls, doors, and windows. These steps are based on heuristic rules applied to the optimized room and furniture layout.

\paragraph{Wall Generation.}
Walls are generated in two stages. First, exterior walls are placed along the boundary of the floor. Second, interior walls are generated at the interface between adjacent but distinct functional areas, such as between two different rooms or between a room and the corridor. This ensures that all enclosed spaces are properly delineated.

\paragraph{Door Placement.}
For each enclosed room, we identify potential locations for doors. The primary candidates are wall segments adjacent to open areas like living rooms, followed by segments adjacent to corridors. This prioritization ensures better circulation and connectivity. From the valid candidate locations—those unobstructed by furniture—a door is randomly placed. Additionally, a main entrance door is added at the designated entry point of the floorplan.

\paragraph{Window Placement.}
Windows are placed on the exterior walls of the floor. For each room and corridor, we identify all segments of its boundary that lie on the floor's exterior and are not blocked by furniture. We then sample from these candidate locations to place several windows, ensuring adequate natural light.

\begin{figure*}[t]
    \centering
    \includegraphics[width=0.95\linewidth]{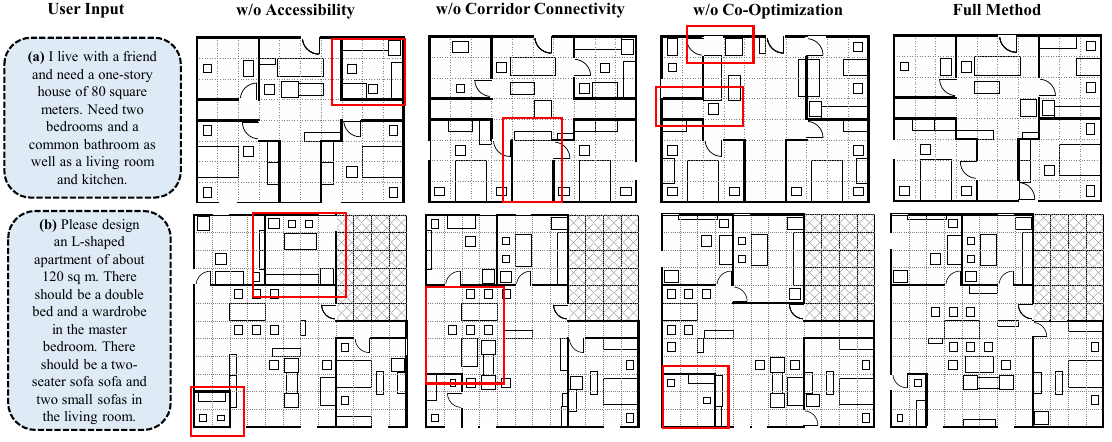}
    \caption{Ablation studies on the impact of key constraints and co-optimization. Outdoor areas are indicated by cross-hatching, walls by thick black lines, doors by sectors, windows by white line segments, and furniture by rectangles. Unreasonable layouts, such as inaccessible areas or rooms, are highlighted with red boxes.}
    \label{fig:ablation_constraints}
\end{figure*}

\section{Additional Ablation Studies}
We evaluate the impact of two key constraints and co-optimization through ablation studies.
\paragraph{Ablation of Key Constraints.} As shown in \cref{fig:ablation_constraints}, we present results under the same input where only accessibility constraints are removed (second column) and only corridor connectivity constraints are omitted (third column), compared with our full method (fifth column).
The visual comparison clearly shows that both constraints are crucial.
Without the accessibility constraint, furniture often blocks potential locations for room entrances, rendering rooms inaccessible.
Similarly, when the corridor connectivity constraint is removed, the resulting layout may contain disjointed corridor segments, making certain areas of the floor unreachable.

\paragraph{Ablation of Co-optimization.} We compare our co-optimization approach with a sequential one that first optimizes rooms and then furniture (fourth column of \cref{fig:ablation_constraints}). This sequential process often results in unreachable rooms and disconnected corridors. Such issues arise because the furniture optimization phase lacks global information and constraints regarding room accessibility and connectivity, which are only considered during the room layout stage.

By modeling the layout using a grid representation and optimizing rooms and furniture jointly through coupled constraints, we explicitly encode connectivity and accessibility constraints and respect them in the final solution.

\begin{figure*}[t]
    \centering
    \includegraphics[width=0.95\linewidth]{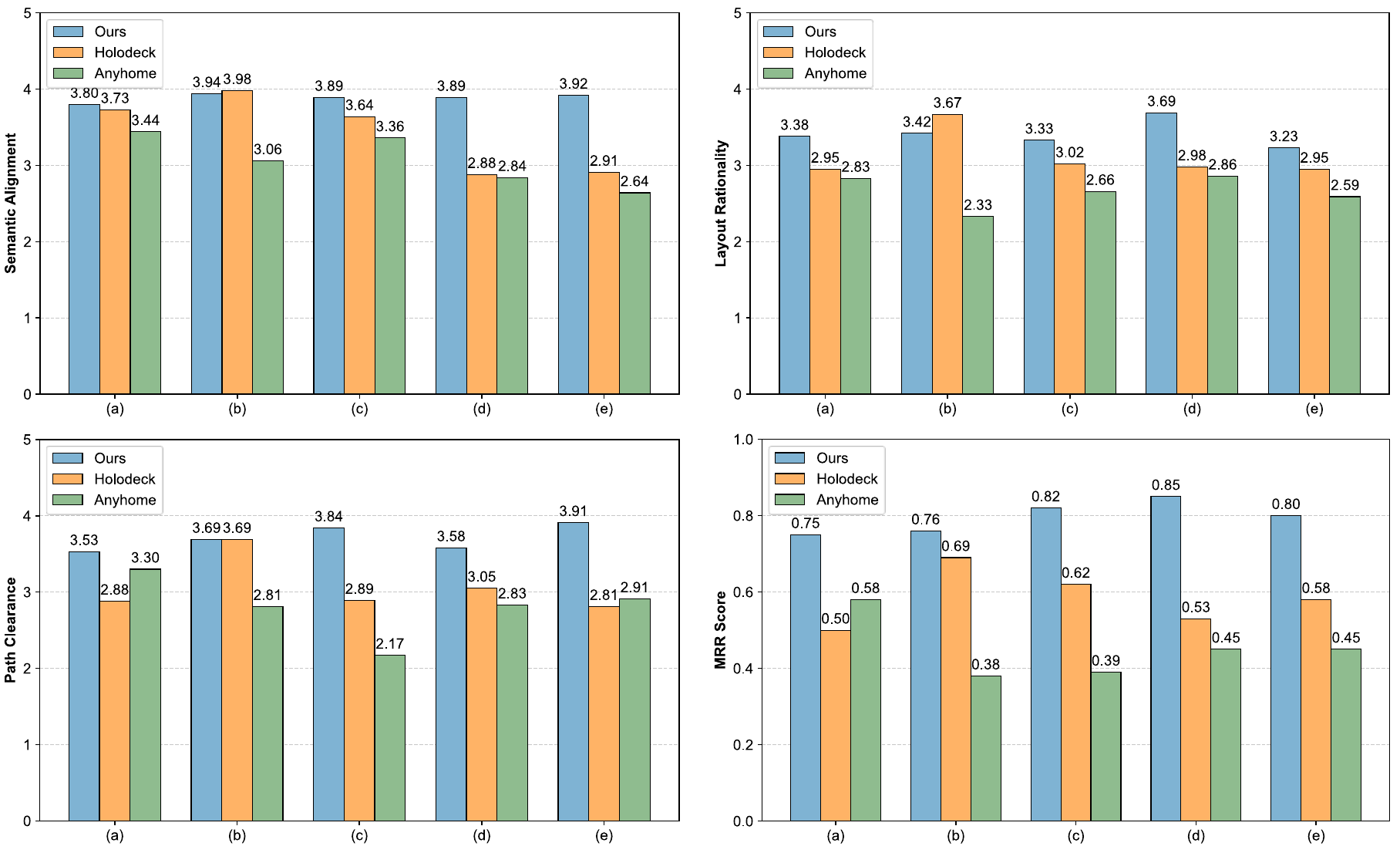}
    \caption{Detailed user study results for each of the five test cases, comparing our method with Holodeck and AnyHome. Our method (blue) consistently outperforms the baselines across all metrics and nearly all cases.}
    \label{fig:user_study_details}
\end{figure*}

\section{Detailed Results for User Study}
As mentioned in the main paper, we conducted a comprehensive user study to evaluate our method against two state-of-the-art baselines: Holodeck and AnyHome. The study involved rating five distinct test cases on three criteria—semantic alignment, layout rationality, and path clearance—and ranking the methods for each case.
\cref{fig:user_study_details} presents the detailed per-case results for each metric. The charts clearly show that our method obtained an overwhelming advantage across all metrics.
Furthermore, our method achieved a significantly higher Mean Reciprocal Rank (MRR), indicating that participants preferred our layouts over the baselines. These detailed results reinforce the aggregated findings in the main paper, highlighting our method's superior performance in generating high-quality, functional, and user-preferred interior layouts.

\end{document}